\documentclass[conference]{IEEEtran}

\usepackage[english]{babel}
\usepackage[utf8]{inputenc}
\usepackage{amsmath}
\usepackage{amsfonts}
\usepackage{amssymb}
\usepackage{graphicx}
\usepackage{algorithm}
\usepackage{algorithmic}

\usepackage{amsmath}
\usepackage{amsthm}
\usepackage{bbm} % for indicator
\usepackage{subfig}
\usepackage{url}

\author{Oren Zeev-Ben-Mordehai, Wouter Duivesteijn, Mykola Pechenizkiy}

\title{Controversy Rules --- Discovering Regions Where Classifiers (Dis-)Agree Exceptionally}
\newtheorem{definition}{Definition}

\begin{document}

%\twocolumn

\maketitle

%%%%%%%%%%%%%%%%%%%%%%%%%%%%%%%
% Abstract                    %
%%%%%%%%%%%%%%%%%%%%%%%%%%%%%%%

\begin{abstract}
Finding regions for which there is higher controversy among different classifiers is insightful with regards to the domain and our models.
Such evaluation can falsify assumptions, assert some, or also, bring to the attention unknown phenomena.
The present work describes an algorithm, which is based on the Exceptional Model Mining framework, and enables that kind of investigations. We explore several public datasets and show the usefulness of this approach in classification tasks. We show in this paper a few interesting observations about those well explored datasets, some of which are general knowledge, and other that as far as we know, were not reported before.
\end{abstract}


%%%%%%%%%%%%%%%%%%%%%%%%%%%%%%%
% Introduction                %
%%%%%%%%%%%%%%%%%%%%%%%%%%%%%%%

\section{Introduction}

In the recent flood of papers analyzing the details of the inner workings of classifiers \cite{Robnik-Sikonja2008,Henelius2014,Duivesteijn2014,Ribeiro:2016:WIT:2939672.2939778}, the attention typically is focused on a single classifier.  We might want to know how a black-box classifier arrives at its predictions \cite{Henelius2014,Ribeiro:2016:WIT:2939672.2939778}, where the classifier predicts well or badly \cite{Duivesteijn2014}, which input attributes influence the output predictions \cite{Robnik-Sikonja2008} and to which degree.  Important as that might be, we propose that more can be learned by investigating the collective behavior of a \emph{set} of classifiers.  Let us illustrate this with a practical example.

Suppose that we work at a bank, and we have to decide on whether or not to lend a mortgage to a series of customers.  We have a rule-based system in place to make this decision.  Since the economic tide ebbs and flows over time, we may need to adapt the rule-based system periodically, to achieve appropriate results.  On every point in time, the system can predict for every customer whether the person gets the mortgage or not.  Interesting would be to find out when and why the rule-based system changes its mind: if subsequent iterations of the system suddenly grants a loan to a previously rejected customer, or vice versa, the era of responsible data science compels us to properly motivate why.  Ideally, we would not just identify single customers for which this holds, but coherent groups of customers that come with a concise description: it would be interesting to know if the system has changed its mind about granting mortgages to people under the age of thirty with at least two kids, for example.  Such descriptions give us more information on whether the behavior displayed by the system is, in fact, desirable.

In this paper we introduce few variants of the following problem. \textbf{Given a dataset, and a collection of relevant classifiers, identify and name the regions of the domain for which there is a high disagreement}.
We describe an algorithm which is based on the Exceptional Model Mining framework~\cite{10.1007/978-3-540-87481-2_1,Duivesteijn2016}, and provide quality measures to address a few possible motivations and preferences. We evaluate the usefulness of the algorithm on publicly available datasets and bring qualitative and quantitative findings.

\section{Related work}
\label{sec:Related}

Given a classifier and a relevant dataset, investigating the interactions among the model and the data is often referred to as model debugging, providing model transparency, or also model interpretability.
Some interpretability mechanisms treat a model as a black box \cite{Henelius2014,10.1007/978-3-319-17091-6_5,Ribeiro:2016:WIT:2939672.2939778,Adler2018} while other employ methods that are tailored to specific classification techniques.
The algorithm described in~\cite{Duivesteijn2014}, enables to investigate a single soft~classifier against a dataset. It requires that the ground truth is provided, and also that the model outputs probabilities (it is a soft classifier). The method investigates the degree to which the ranking of the model in a specific subgroup is in agreement with the ground truth. This is done by counting the obvious errors (when a negative is ranked before a positive). Regions for which the rate of obvious errors is significantly higher, or significantly lower, from the same measure for the whole dataset, are then reported.

Black box auditing, or discrimination aware approach, is described in~\cite{Adler2018}.
\textsf{GoldenEye}/\textsf{GoldenEye++}~\cite{Henelius2014,10.1007/978-3-319-17091-6_5} is highlighting the feature importance/feature interaction by shuffling the values in columns of specific features within a predicted label, and measuring the label changes.
\textsf{EXPLAIN}~\cite{Robnik-Sikonja2008,Robnik-Sikonja2012} checks the effect of blinding the model with respect to values of a specific attribute.
\textsf{SHAP}~\cite{2018arXiv180203888L} divides the contribution to a classification among the features of a case.
\textsf{LIME}~\cite{Ribeiro:2016:WIT:2939672.2939778} attempts to describe the model in a locality of a case under scrutiny using an interpretable proxy.
Also interesting approaches for interpretability explore regions of uncertainty~\cite{Gal2016Uncertainty}, attention given a case under test~\cite{selvaraju2016grad}, or cognitive psychology traits of the model~\cite{Ritter2017}.
Some works extract rules, or provide a simplified model~\cite{2017arXiv170101358L,lakkaraju2016interpretable,Hara2016}.

If one manages to compress the model and the data, for example following the \emph{Minimum Description Length} (MDL) framework~\cite{Cook1994}, then they somehow capture the essence of the model/data. One should describe a model and then the exceptions in the data that do not follow the model and to find the point for which the overall description is minimal (the model is described in sufficient detail and the leftover exceptions are few).

\section{Prerequisites}
\label{sec:Prerequisites}

Assume given a dataset $DS$ from a domain $\mathcal{D}$, consisting of $m$ \emph{cases} (or \emph{records} of the form $r=\left(\text{attr}_1,\text{attr}_2,\ldots,\text{attr}_k,\ell\right)
$. 
We refer to the final element of each case, as \emph{target}, or also the \emph{true label}. The target is nominal, with the set of possible values $\mathcal{C}$, thus $\ell \in \mathcal{C}$. All other elements of each case are referred to as the \emph{attributes}, which can be either of a numeric type, or a nominal type. While the domain of each individual attribute is left free, we denote the collective domain of the $k$ attributes by $\mathcal{A}$. This notation allows us to formally define what a classifier is:

\begin{definition}[Classifier]
Given a domain $\mathcal{D}$ with collective attribute domain $\mathcal{A}$ and the nominal type of the target $\mathcal{C}$, a classifier $C$ is a function $C: \mathcal{A} \to \mathcal{C}$, assigning a label to every possible input value from $\mathcal{A}$.
\end{definition}

The main goal of a classifier, as it is generally understood in machine learning, is to predict: assigning labels to cases whose real target value we do not know. To arrive at a formal definition of such predictions, we need to introduce some more notation. Let $DS$ be a dataset, where the true labels, in the general case, are not known.
We denote by superscript $i$ the $i$\textsuperscript{th} case of the dataset $DS$ or elements thereof. Hence, the first case is denoted by $r^1$, the target value of the seventh case, whether it is known or not, by $\ell^7$, and the value for the fourth attribute in the eighth case by $\text{attr}_4^8$.

\begin{definition}[Predictions]
Given a dataset $DS$ consisting of $m$ cases, and a classifier $C$, we define the \emph{predictions} of $C$ on $DS$ to be the vector $(y_C^1,\ldots,y_C^m) \in \mathcal{C}^m$, where $y_C^i = C(\text{attr}_1^i,\ldots,\text{attr}_k^i)$.
\end{definition}

Hence, the vector of predictions collects the outputs of the classifier function $C$ on all cases in the dataset $DS$.

The main goal of this paper is to find regions, or subgroups of cases, of high controversy across a set of classifiers. Hence, we assume as given a set of $n$ classifiers $\{C_1,\ldots,C_n\}$. For the purposes of this paper, it is irrelevant exactly how any of these classifiers arrive at their predictions: we are agnostic of the internal workings of a classifier function. Instead, we merely analyze them in terms of their predictions:

\begin{definition}[Prediction matrix]
Given a dataset $DS$ consisting of $m$ cases, and a set of $n$ classifiers $\{C_1,\ldots,C_n\}$, the \emph{prediction matrix} $M$ is the $(m\times n)$-matrix with entries from $\mathcal{C}$ defined by:
$$
M_{ij} = y_{C_j}^i
$$
\end{definition}

Hence, the first row of the prediction matrix $M$ collects the predictions of all $n$ classifiers for the first case in the original matrix $DS$, etcetera.

\subsection{Local Pattern Mining}

We would like to identify one or more subgroups of the cases, for example $SG \subset DS$, such that on average for cases $i \in SG$, and the $n$ classifiers $j \in (1 \dots n)$, there is high controversy among the relevant entries $M_{ij}$. 
%As a side note, if we are given the prediction matrix $M$, for example as a flat file, or as a database table, we don't need the original classifiers to proceed, the predictions matrix $M$ suffices.
The ground truth, $GT$, for the classification problem, referred above as the target values, or the true labels, is not a necessity for the problem we describe next in its basic form, yet once present, new options and questions can be investigated.

When selecting a subset of the cases in $DS$, we restrict ourselves to regions that can be identified with a description $desc$ that belongs to a description language $\mathcal{L}$. Thus for example, if $desc_1 \equiv \text{Class} = 3 \land \text{Age} < 5$ is a valid description in $\mathcal{L}$, then the matching subgroup of cases $SG_{desc_1}$, those for which the description $desc_1$ evaluates to true, is a valid candidate as a subgroup.
This is often the approach with Subgroup Discovery~\cite{Klosgen:2002:HDM:778212}, and with Exceptional Model Mining~\cite{10.1007/978-3-540-87481-2_1,Duivesteijn2016}.
In \emph{Subgroup Discovery} (SD), one can identify the most interesting subgroups w.r.t.\ a single target. With \emph{Exceptional Model Mining} (EMM), one can address multiple target attributes when evaluating how exceptional a subgroup is. Both frameworks require that one declares a set of attributes that can be part of the \emph{description} for a subgroup, therefore, the identification of the region. Also required is a single or, for EMM, a set of attributes, that are used when evaluating the exceptionality of the region.
Formally both SD, and EMM, require a declaration of a subset of the attributes of the dataset, $\{a_1,\dots,a_k,t_1,\dots,t_m\}$, where $\{a_1,\dots,a_k\}$ are used to describe subgroups, and $\{t_1,\dots,t_m\}$ are used to evaluate subgroups. Thus the description language $\mathcal{L}$ is based on $\{a_1,\dots,a_k\}$ and the relevant domains. Given a dataset $DS$, a description, $desc \in \mathcal{L}$, is interchangeable with the subgroup $SG_{desc} \subset DS$ that corresponds to the cases $i \in DS$ for which $desc(a_1^i,\dots,a_k^i)$ is true. For evaluating the subgroups, as mentioned above, $\{t_1,\dots,t_m\}$ are used. For SD, $m = 1$, for EMM, $m \geq 1$.  
Of course there are many ways to evaluate the exceptionality of a subgroup, for EMM in particular, but also for SD. Therefore a specific \emph{quality measure} $\varphi : \mathcal{L} \rightarrow \mathbb{R}$ (for EMM, based on a \emph{model class}) must be chosen to evaluate the quality of the region in terms of exceptionality. 
Hence, $\varphi(desc)$ assigns a value to the description $desc$ based on the $\{t_1,\dots,t_m\}$ attributes of the relevant entries in $SG_{desc}$.
A reasonable choice to realize the search involved with SD or EMM, is with the \emph{Beam Search} algorithm~\cite{BeamSearch}.

\section{The Controversy Rules Model Class for EMM}
\label{quality measures}

Our prerequisites and the standard EMM terminology can be naturally mapped onto one another, as follows.  The descriptors $\{a_1,\dots,a_k\}$ from EMM will be $\{\text{attr}_1,\text{attr}_2,\ldots,\text{attr}_k\}$ of the dataset, and the targets $\{t_1,\dots,t_m\}$ of EMM will be the $n$ predictions from $M$.  In some situations we augment $M$, where available and relevant, with the ground truth label $\ell$.

%We next describe what we believe to be an interesting questions with respect to the collection of classifiers and the controversy over cases in $DS$, as can be evaluated against $M$ during the search over $DS$ and the possible descriptions given the collective attribute domain $\mathcal{A}$. 
We illustrate the core concept of Controversy Rules by a single case, or row, $r^1$ compared to another row $r^2$.
If the set of $n$ predictions over $r^1$ has higher entropy than the set over $r^2$, we would claim that $r^1$ is more interesting than $r^2$. We use here the base 2 Shannon entropy, $\mathcal{H}:\mathcal{C}^n\rightarrow[0,\infty)$:
\begin{align*}
\mathcal{H}(c_1,\ldots,c_n)=-\sum\limits_{c \in \mathcal{C}} P(c)\cdot\log_2 P(c)\\
\text{where for every } c \in \mathcal{C}, P(c) = \dfrac{1}{n}\left(\sum\limits_{i=1}^n \mathbbm{1}(c_i = c)\right)
\end{align*}
Following this definition, if we have $10$ classifiers ($n=10$), and a binary target, then a row where five classifiers predict the one label and five the other, is more interesting than if the votes were six versus four.  If we have $3$ classes, then a tally of $(3,4,3)$ is as interesting as of $(4,3,3)$ and both are more interesting than a tally of $(1,1,8)$. This is of course if we look for regions with disagreement. If we seek for regions with high agreement, we prefer the lower entropy.

Now consider subgroups of the cases, or collections of rows. As to help the reader to follow the intuition, we give two $M$ matrices for two toy datasets and their respective classifiers, in Table~\ref{tab:sample1} and Table~\ref{tab:sample2}. For simplicity, assume that a description exists for each subset of both toy datasets. Therefore one can name any of those subsets and evaluate their quality measures. Below we refer, for example, to the subgroup containing rows 1 and 3 as subgroup $\{1,3\}$.

\begin{table}[t]
\centering
\makebox[0pt][c]{\parbox{0.5\textwidth}{%
    \begin{minipage}[b]{0.49\hsize}\centering
        \caption{Toy dataset $A$}
        \label{tab:sample1}
		\begin{tabular}{|r|c|c|c|c|l|}
		\hline
		& $C_1$ & $C_2$ & $C_3$ & $C_4$ & \\ 
		\hline
		1. & 1 & 1 & 0 & 1 & \\ 
		\hline
		2. & 0 & 1 & 0 & 1 & * \\ 
		\hline
		3. & 1 & 0 & 1 & 1 & \\ 
		\hline
		4. & 1 & 1 & 1 & 0 & \\ 
		\hline
		5. & 0 & 1 & 1 & 0 & * \\ 
		\hline
		6. & 0 & 0 & 0 & 0 & \\ 
		\hline
		7. & 0 & 0 & 0 & 0 & \\ 
		\hline
		8. & 0 & 0 & 0 & 1 & \\ 
		\hline 
		\end{tabular}
    \end{minipage}
    \hfill
    \begin{minipage}[b]{0.49\hsize}\centering
        \caption{Toy dataset $B$}
        \label{tab:sample2}
		\begin{tabular}{|r|c|c|c|c|l|}
		\hline
		& $C_1$ & $C_2$ & $C_3$ & $C_4$ & \\ 
		\hline
		1. & 1 & 1 & 0 & 1 & * \\ 
		\hline
		2. & 0 & 1 & 0 & 1 & \\ 
		\hline
		3. & 1 & 1 & 0 & 1 & * \\ 
		\hline
		4. & 1 & 1 & 0 & 1 & * \\ 
		\hline
		5. & 0 & 1 & 1 & 0 & \\ 
		\hline
		6. & 0 & 0 & 0 & 0 & \\ 
		\hline
		7. & 0 & 0 & 0 & 0 & \\ 
		\hline
		8. & 0 & 0 & 0 & 1 & \\ 
		\hline 
		\end{tabular}
    \end{minipage}
    \hfill
}}
\end{table}

\subsection{Row Controversy}
% \varphi_{row}
In the first scenario, we seek regions with \emph{high per row controversy} across the classifiers. We measure this by mean per row entropy over the cases in a subgroup. Therefore the quality measure that we use here is: $$
\varphi_{row}(SG) \equiv\\
\dfrac{1}{|SG|}\sum\limits_{i \in SG}\mathcal{H}(M_{i1},\dots,M_{in})
$$
Note that we ignore the identity of the classifiers, or the actual predictions, and we just evaluate the mean per row entropy for the subgroup. 
We set a minimum threshold for number of rows, so that the reported subgroups are actionable, yet other from that, a smaller subgroup with higher mean entropy is still ranked before bigger subgroups with smaller mean entropy. The use case for this scenario is when we are interested in subgroups of the domain for which different classifiers predict differently or even completely at random. The rationale for this desire, \emph{described here, for simplicity, in binary classification terms}, is that we are less concerned by a big subgroup, for which at any given row, one classifier gets it wrong (or only one gets it right), while the other get it right (wrong), than by a smaller subgroup for which always half of the classifiers get those cases wrong. The subgroup on which half of the classifiers get the cases wrong should be ranked higher. In Table~\ref{tab:sample1}, toy dataset $A$ and its relevant classifiers, we would like to discover first the subgroup $\{2,5\}$.

\subsection{Consistent Classification}
% \varphi_{consistent\_classification}
In the next scenario, we consider the following objective. We are interested in controversy but of less random nature: a scenario in which few classifiers consistently differ from the other classifiers. \emph{We assume here that the classifiers are consistent in the regions (low entropy per classifier)}. In the example from Table~\ref{tab:sample2}, toy dataset $B$, we would like to discover first subgroup $\{1,3,4\}$ or subgroup $\{1,2,3,4\}$. This is because all four classifiers, $\{C_1,C_2,C_3,C_4\}$ are each internally consistent in those regions, while there is a disagreement across the four. Notice that using that intuition, \emph{we direct the search to a region in which the per-classifier entropy is low}, but mean per-row entropy is high. To this end, we define the following quality measure:
\begin{align*}
\varphi_{ccl}(SG) &\equiv \dfrac{1}{|SG|}\sum\limits_{i \in SG}\mathcal{H}(M_{i1},\dots,M_{in})\\
& - \dfrac{1}{n}\sum\limits_{j=1}^n \mathcal{H}(M_{ij} | i \in SG)
\end{align*}
For example, $\varphi_{ccl}(\{1,2,3,4\})=0.858 - 0.203 = 0.656$ (rounded), $\varphi_{ccl}(\{1,3,4\})=0.811 - 0 = 0.811$, $\varphi_{ccl}(\{5,6,7,8\})=0.453 - 0.608 = -0.156$, $\varphi_{ccl}(\{5,6,7\})=0.333 - 0.459 = -0.126$, and $\varphi_{ccl}(\{5,6\})=0.5 - 0.5 = 0$. 
The use case for this scenario is to identify regions in which few classifiers behave different, yet limiting the search for regions in which each classifier is consistent.
%, therefore not necessarily discovering the existence of classifiers that are the negation of other classifiers.

\subsection{Consistent Accordance}
% \varphi_{consistent\_accordance}
We next identify controversy of consistent nature, while overcoming the rigidity of $\varphi_{ccl}$, where different predictions over different cases result in high classifier-wise entropy, thus lower rank for the relevant subgroup.
Achieving this goal allows us to identify regions where a few classifiers are the negation of the majority. We cannot normally achieve this with $\varphi_{row}$, unless the same regions indeed contain the greatest per-row entropy on average. To allow for different predictions per-classifier we move from the prediction space to the accordance space. Thus we first identify the top predicted class per row (most frequently predicted), and then compare it to the prediction. In case of a tie, we choose one of the classes. Thus for every row, $i$, $Top_i \leftarrow most\_frequent\_in\_row(M_i)$, and then for every classifier $j$, $M'_{ij} \leftarrow \mathbbm{1}(M_{ij} = Top_i)$. We next search for interesting regions based on $M'$, using the following quality measure:
\begin{align*}
\varphi_{cac}(SG) &\equiv \dfrac{1}{|SG|}\sum\limits_{i \in SG}\mathcal{H}(M'_{i1},\dots,M'_{in})\\
& - \dfrac{1}{n}\sum\limits_{j=1}^n \mathcal{H}(M'_{ij} | i \in SG)
\end{align*}

\subsection{Consistent Correctness}
% \varphi_{consistent\_correctenss}
In this scenario and all subsequent ones, we assume the availability of the ground truth, $GT$.
The availability of the ground truth enables us to attempt to identify hard-to-classify regions, on which few models actually succeed, or the other way around: easy regions, on which a few models consistently fail. We start by collecting the correctness of the predictions, hence for every case $i$ and for every classifier $j$, $M''_{ij} \leftarrow \mathbbm{1}(M_{ij} = \ell_i)$. We then evaluate using the mean of row-wise entropies minus the mean of classifier-wise entropies. Note that also here, once we switch from the output space to the correctness space, the per classifier consistency is of a different nature. Hence, classifiers that are the negation of other classifiers may result in higher ranking for the relevant regions. We use the following quality measure:
\begin{align*}
\varphi_{cco}(SG) &\equiv \dfrac{1}{|SG|}\sum\limits_{i \in SG}\mathcal{H}(M''_{i1},\dots,M''_{in})\\
& - \dfrac{1}{n}\sum\limits_{j=1}^n \mathcal{H}(M''_{ij} | i \in SG)
\end{align*}
Differences between $\varphi_{cco}$ and $\varphi_{cac}$ are possible, where there are cases for which the majority of classification is different from the true label.

\subsection{Ground Truth as Yet Another Classifier}
% \varphi_{GT\_as\_yac}
%Here we also assume the availability of $GT$.
%We want to identify regions that are hard to classify.
%One idea that comes into mind, is to count the errors per row, and to aggregate those for the region. However the exact aggregation needs to be carefully thought, should a simple mean be the right design choice, or maybe mean of the squares.
%We suggest an alternative using disagreement among the classifiers. 
If we treat the ground truth as yet another classifier, we can evaluate the mean per-row entropy as is done for $\varphi_{row}$.
What is the effect of adding $GT$ as an additional classifier? Rows for which most of the classifiers predict correctly, now have a lower entropy. Rows for which only a minority of the classifiers predict rightly, have a higher entropy. The search for regions for which the mean row-wise entropy is the highest, results in finding regions that are hard to predict correctly.
We add $GT$ as an additional classifier, as described above, and also in another experiment, add $GT$ as additional $n$ classifiers.
\begin{align*}\varphi_{GT\_as\_yac}(SG) &\equiv
\dfrac{1}{|SG|}\sum\limits_{i \in SG}\mathcal{H}(M_{i1},\dots,M_{in},\ell^i)\\
\varphi_{GT\_as\_yac'}(SG) &\equiv
\dfrac{1}{|SG|}\sum\limits_{i \in SG}\mathcal{H}(M_{i1},\dots,M_{in},{(\ell^i)}_{\times n})
\end{align*}
Note that $\varphi_{GT\_as\_yac}$ is expected to be similar to $\varphi_{row}$ yet puts some additional emphasis on regions with errors. $\varphi_{GT\_as\_yac'}$ should take this aspect even further: by matching each classifier's prediction with a copy of the ground truth, the weight of mistakes, as is reflected in the ranking of the subgroups, should be even higher.

%We opt not to cross these experiments, ground truth as yet another classifier, with the experiments above for consistency, as this does not seem to us to be an interesting evaluation.

\subsection{Relative Average Subranking Loss}
% \varphi_{rasl} (SCaPE)
The last scenario in this paper %, also assumes the presence of the ground truth, $GT$, and is suitable for binary classification. 
is applicable to binary classification only. 
We adapt the existing SCaPE model class for EMM \cite{Duivesteijn2014}, to
identify regions that are exceptionally hard or easy to predict. 
By examining $M$, we calculate the empirical probability of predicting the positive class per row. Thus for every row $i$, $$Prob^i \leftarrow count\_of\_positives\_in\_row(M_i) / n\ .$$
We obtain therefore a soft~classifier, $Prob$, to be contrasted with the ground truth $GT$, gauged with the quality measure $\varphi_{rasl}$ used in SCaPE.

\section{Experiments}
\label{sec:Experiments}

We illustrate the workings of the Controversy Rules model class for EMM, by experimenting on the following classifiers: Decision Tree~\cite{Quinlan:1993:CPM:152181}, Na\"ive Bayes~\cite{nb}, 3-Nearest Neighbors~\cite{doi:10.1080/00031305.1992.10475879}, Random Forest~\cite{TinKamHo1995}, and Support Vector Machine with linear kernel~\cite{Cortes1995}.
The choice of classifiers is purely for illustrative purposes and should not be confused with the core contribution of this paper: we provide a method to find regions of controversy between classifiers, which we illustrate with this selection of well-known classifiers (which should not be taken as endorsement of the classifiers themselves).
We obtain the predictions by running 10-fold cross validation for each of the model classes. Hence, technically, each prediction column is created by 10 different classifiers; the perceived classifiers are virtual, and have never existed. We mention this for the benefit or reproducibility; how the predictions were obtained is not fundamental to the core contribution of this paper.

We run the experiments on the eight datasets listed in Table~\ref{tab:datasets}.  Most are taken from the UCI ML repository~\cite{Dua:2017}. The \emph{Titanic} dataset is taken from Kaggle~(\url{https://www.kaggle.com/c/titanic/data}), and \emph{Pima-indians} (which is no longer available in the UCI ML repository) can also be accessed there~(\url{https://www.kaggle.com/uciml/pima-indians-diabetes-database/data}). 
The \emph{YearPredictionMSD} dataset comes with a naturally in-built regression task (predicting the year in which a song was released). We define our own classification task on this dataset, converting the year into decades (the floor of the year divided by 10 is taken as the true label). Some of the datasets are suitable for binary classification tasks (\emph{Mushroom}, \emph{Titanic}, \emph{Adult}, \emph{Pima-indians}), while other contain more than 2 classes, although sometimes ordinal in nature (\emph{Balance-scale}, \emph{Car}, \emph{YearPredictionMSD}).

To discover subgroups, we employ the Beam Search algorithm for Exceptional Model Mining, as described in \cite[Algorithm 1]{Duivesteijn2016}.  The parameters are set as follows: beam width $w=25$, search depth $d=3$.
To avoid tiny subgroups, we require a minimum support of $4\%$ of the cases.

\begin{table}[t]
\centering
\caption{Datasets used for experiments.}
\label{tab:datasets}
\begin{tabular}{rl|rrrr}
 & Dataset & \#cases & \multicolumn{2}{c}{\#attributes} & $|\mathcal{C}|$ \\
 &         &   ($m$) & discrete & numeric & \\
\hline
1. & Mushroom & 8,124 & 22 & 0 & 2 \\
2. & Titanic & 891 & 3 & 4 & 2 \\
3. & Adult & 48,842 & 8 & 6 & 2 \\
4. & Balance-scale & 625 & 0 & 4 & 3 \\
5. & Car & 1,728 & 6 & 0 & 4 \\
6. & Pima-indians & 768 & 0 & 8 & 2 \\
7. & Covertype & 581,012 & 44 & 10 & 7 \\
8. & YearPredictionMSD & 515,345 & 0 & 90 & 10
\end{tabular}
\end{table}

%%%%%%%%%%%%%%%%%%%%%%%%%%%%%%%
% Mushroom                    %
%%%%%%%%%%%%%%%%%%%%%%%%%%%%%%%

\subsection{Mushroom}
\label{subsec:mushroom}

\begin{figure}[t]
\centering
\includegraphics[width=8cm]{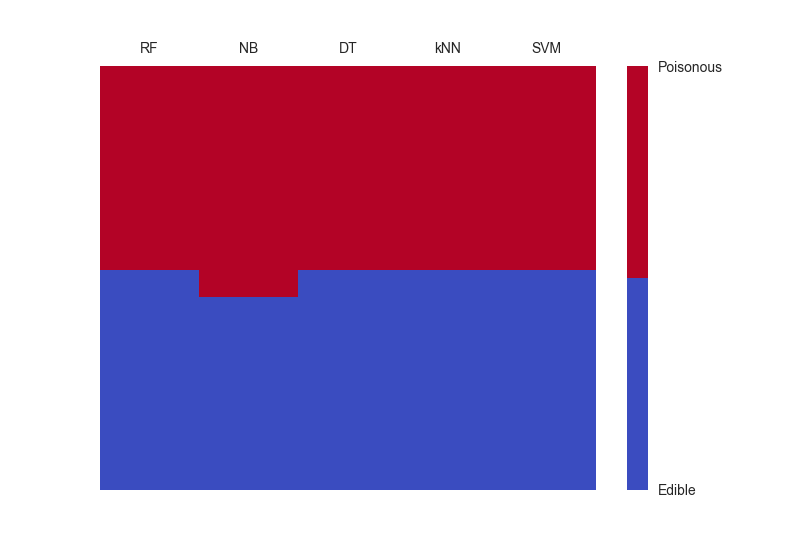}
\caption{Predictions matrix, ordered from left to right based on the classifiers' predictions. The \emph{Mushroom} dataset.}
\label{img:mushroom:fig_predictions_matrix_order}
\end{figure}

On the \emph{Mushroom} dataset, four out of five classifiers predicted almost all test cases correct (Na\"ive Bayes has 216 false positives, and 3 false negatives, k-Nearest Neighbors has 2 false negatives, and the other three classifiers do not make errors).
The prediction matrix is displayed in Figure~\ref{img:mushroom:fig_predictions_matrix_order}.
The order of the classifiers, from left to right, is as those are listed for the experiment. The cases, or the rows, are ordered by the predictions of the classifiers, lexicographically from left to right.
As can be seen, one classifier (Na\"ive Bayes in this case) is predicting differently from the rest for numerous cases, while the other agree almost always.
%We expect that subgroups that are reported by the $\varphi_{row}$ quality measure to contain a lot of those cases where there is controversy, and indeed this is what we see in Figure~\ref{img:mushroom:fig_distribution_of_row_entropies_qmrow_1}. The figure shows the counts of rows per calculated entropy value, and highlights in red are the rows that match the top description found with $\varphi_{row}$ (listed in Table~\ref{tab:mushroom_row}).
The exact descriptions ordered the same, are reported also by $\varphi_{GT\_yac}$ and by $\varphi_{GT\_yac'}$. This is expected as errors and disagreements here are in the same cases.

% Mushroom fig_predictions_matrix_order

% Mushroom fig_distribution_of_row_entropies_qmrow_1
%\begin{figure}[t]
%\centering
%\includegraphics[width=8cm]{Images/Mushroom/fig_distribution_of_row_entropies_qmrow_1}
%\caption{Distribution of per-row entropies, and a specific subgroup highlighted in red. The relevant description for the subgroup is $odor = n \land stalk\_color\_above\_ring \neq g \land stalk\_color\_below\_ring = p$. The \emph{Mushroom} dataset.}
%\label{img:mushroom:fig_distribution_of_row_entropies_qmrow_1}
%\end{figure}

% tab:mushroom_row
\begin{table}[t]
\centering
\caption{Subgroups found with the $\varphi_{row}$ quality measure, for the \emph{Mushroom} dataset ($\varphi_{row}(DS)=0.046$).}
\begin{tabular}{p{6cm} r r}
description & \#cases & $\varphi_{row}$ \\
\hline
$odor = n~\land stalk\_color\_above\_ring \neq g \land stalk\_color\_below\_ring = p$ & 384 & 0.385 \\
$ring\_type = p \land stalk\_color\_above\_ring \neq g \land stalk\_color\_below\_ring = p$ & 384 & 0.385 \\
$bruises = t \land stalk\_color\_above\_ring \neq g \land stalk\_color\_below\_ring = p$ & 384 & 0.385 \\
$odor = n \land stalk\_color\_above\_ring = p \land stalk\_color\_below\_ring \neq g$ & 384 & 0.376 \\
$ring\_type = p \land stalk\_color\_above\_ring = p \land stalk\_color\_below\_ring \neq g$ & 384 & 0.376 \\
$bruises = t \land stalk\_color\_above\_ring = p \land stalk\_color\_below\_ring \neq g$ & 384 & 0.376 \\
\end{tabular}
\label{tab:mushroom_row}
\end{table}

% tab:mushroom_ccl
\begin{table}[t]
\centering
\caption{Subgroups found with the $\varphi_{ccl}$ quality measure, for the \emph{Mushroom} dataset ($\varphi_{ccl}(DS)=-0.952$).}
\begin{tabular}{p{6cm} r r}
description & \#cases & $\varphi_{ccl}$ \\
\hline
$odor = n \land stalk\_color\_above\_ring \neq g \land stalk\_color\_below\_ring = p$ & 384 & 0.186 \\
$odor = n \land stalk\_color\_above\_ring = p \land stalk\_color\_below\_ring \neq g$ & 384 & 0.176 \\
$gill\_color \neq u \land odor = n \land stalk\_color\_below\_ring = p$ & 432 & 0.143 \\
$gill\_color \neq u \land odor = n \land stalk\_color\_above\_ring = p$ & 432 & 0.135 \\
$gill\_color \neq n \land odor = n \land stalk\_color\_below\_ring = p$ & 432 & 0.114 \\
$gill\_color = p \land gill\_spacing = c \land odor = n$ & 468 & 0.110 \\
\end{tabular}
\label{tab:mushroom_ccl}
\end{table}

\begin{figure}[t]%
    \centering
    \subfloat[Top description both for $\varphi_{row}$ and for $\varphi_{ccl}$.	$odor = n \land stalk\_color\_above\_ring \neq g \land stalk\_color\_below\_ring = p$\label{fig:subrow}]{{\includegraphics[width=8cm]{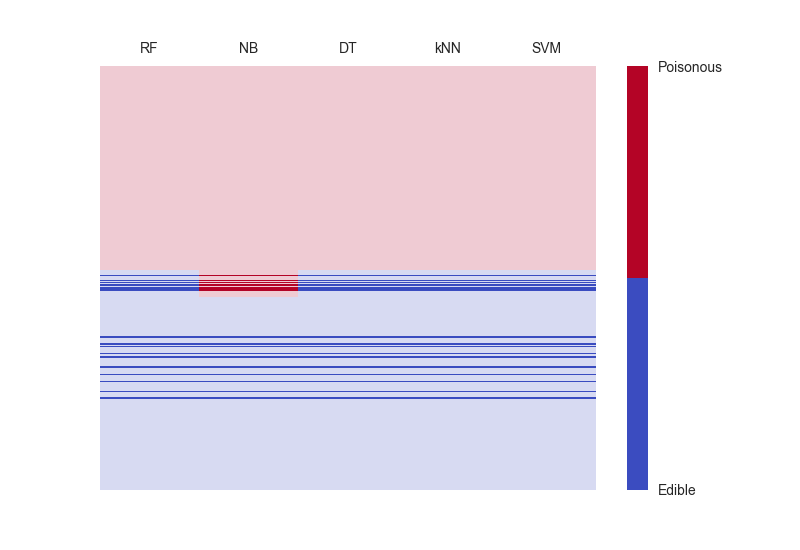}
    }}%
    \qquad
    \subfloat[A new description reported by $\varphi_{ccl}$. $gill\_color \neq u \land odor = n \land stalk\_color\_below\_ring = p$\label{fig:subccl}]{{\includegraphics[width=8cm]{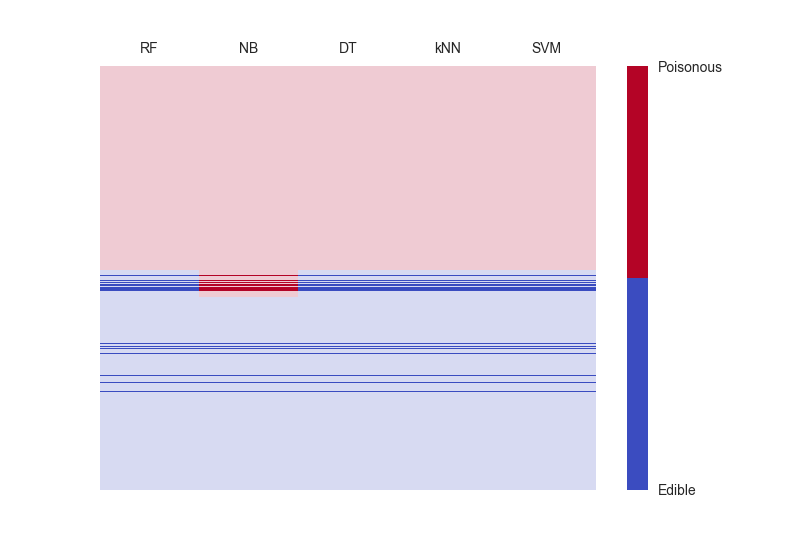} 
    }}%
    \caption{Prediction matrices on the \emph{Mushroom} dataset, with highlighted subgroups.}%
    \label{img:mushroom:compare_row_ccl}
\end{figure}

% tab:mushroom_cac
\begin{table}[t]
\centering
\caption{Subgroups found with the $\varphi_{cac}$ quality measure, for the \emph{Mushroom} dataset ($\varphi_{cac}(DS)=-0.023$).}
\begin{tabular}{p{6cm} r r}
description & \#cases & $\varphi_{cac}$ \\
\hline
$gill\_color = w \land habitat = d \land stalk\_color\_above\_ring \neq g$ & 390 & 0.109 \\
$gill\_color = w \land habitat = d \land stalk\_color\_below\_ring \neq g$ & 390 & 0.109 \\
$gill\_color = w \land odor = n \land population = v$ & 342 & 0.096 \\
$gill\_color = w \land stalk\_color\_below\_ring \neq w \land stalk\_root = b$ & 352 & 0.092 \\
$gill\_color = w \land stalk\_color\_above\_ring \neq g \land stalk\_root = b$ & 544 & 0.090 \\
$gill\_color = w \land stalk\_color\_below\_ring \neq g \land stalk\_root = b$ & 544 & 0.090 \\
\end{tabular}
\label{tab:mushroom_cac}
\end{table}

% Mushroom fig_predictions_matrix_order=True_qmcac_1
\begin{figure}[t]
\centering
\includegraphics[width=8cm]{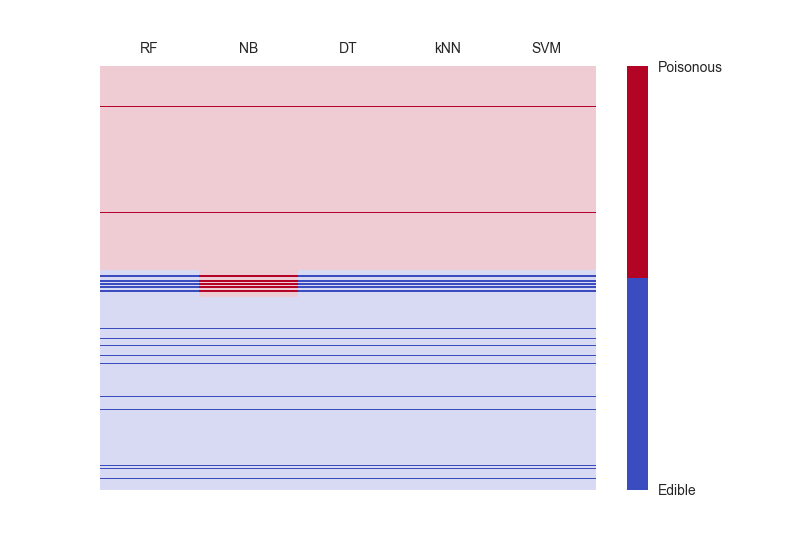}
\caption{Top subgroup found with $\varphi_{cac}$ on the \emph{Mushroom} dataset: $gill\_color = w \land habitat = d \land stalk\_color\_above\_ring \neq g$.}
\label{img:mushroom:fig_predictions_matrix_order=True_qmcac_1}
\end{figure}

$\varphi_{rasl}$ did not find anything interesting as the errors made (by the Na\"ive Bayes classifier) are to mistake consistently a negative to be a positive for a few of the cases, and hence the probability for those cases is indeed between the 0 for most of the negative cases, and the 1 for most of the positive cases.

Table~\ref{tab:mushroom_ccl} lists the descriptions reported by the $\varphi_{ccl}$ quality measure.  The top subgroup is the same as the one found with $\varphi_{row}$, but subsequent subgroups differ.  To illustrate the difference, Figure~\ref{img:mushroom:compare_row_ccl} displays two prediction matrices: one (Figure \ref{fig:subrow}) for the top subgroup for both measures, and one (Figure \ref{fig:subccl}) for the new description $gill\_color \neq u \land odor = n \land stalk\_color\_below\_ring = p$. Comparing those two subgroups, we note that as one classifier, the Na\"ive Bayes, is more consistent when there are fewer negative cases, relevant descriptions are being ranked higher with $\varphi_{ccl}$.    

Table~\ref{tab:mushroom_cac} lists the subgroups found with $\varphi_{cac}$. The top description $gill\_color = w \land habitat = d \land stalk\_color\_above\_ring \neq g$ is illustrated in Figure~\ref{img:mushroom:fig_predictions_matrix_order=True_qmcac_1}. We see that some more positive cases are included. The internal accordance of the classifiers is intact by adding those cases, and this subgroup is more interesting than the top one reported by $\varphi_{row}$, if taking into account also the consistent accordance. The descriptions reported by $\varphi_{cco}$ are the same as those reported by $\varphi_{cac}$. This is not surprising, since the majority of the classifiers get all the cases correct.

%%%%%%%%%%%%%%%%%%%%%%%%%%%%%%%
% Balance                     %
%%%%%%%%%%%%%%%%%%%%%%%%%%%%%%%

\subsection{Balance-scale}

% Balance fig_distribution_and_sg_qmrow_1
%\begin{figure}[t]
%\centering
%\includegraphics[width=8cm]{Images/Balance/fig_distribution_and_sg_qmrow_1}
%\caption{Distribution of values and true labels, the \emph{Balance-scale} dataset. Highlighted in red is the subgroup matching the description $Left\_Weight \le 2 \land Right\_Distance \le 2 \land Right\_Weight > 3$. This description is ranked highest by the $\varphi_{row}$ quality measure.}
%\label{img:balance:fig_distribution_and_sg_qmrow_1}
%\end{figure}

% Balance fig_distribution_of_predictions
%\begin{figure}[t]
%\centering
%\includegraphics[width=8cm]{Images/Balance/fig_distribution_of_predictions}
%\caption{Distribution of predictions, the \emph{Balance-scale} dataset}
%\label{img:balance:fig_distribution_of_predictions}
%\end{figure}

% Balance fig_predictions_correlation
%\begin{figure}[t]
%\centering
%\includegraphics[width=8cm]{Images/Balance/fig_predictions_correlation}
%\caption{Correlation among the classifiers' predictions, \emph{Balance-scale} dataset}
%\label{img:balance:fig_predictions_correlation}
%\end{figure}

% Balance fig_parallel_coordinates
\begin{figure}[t]
\centering
\includegraphics[width=8cm]{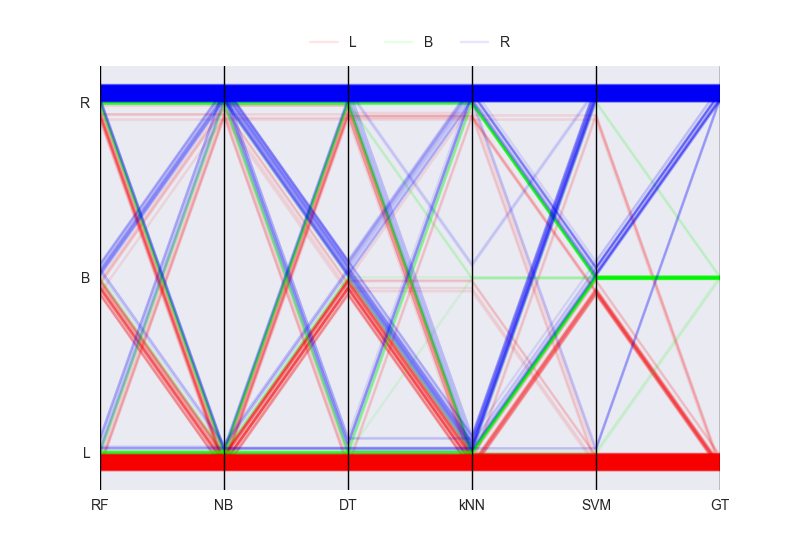}
\caption{Parallel coordinates for the predictions and the true labels, \emph{Balance-scale} dataset}
\label{img:balance:fig_parallel_coordinates}
\end{figure}

% tab:balance_row
\begin{table}[t]
\centering
\caption{Subgroups found with the $\varphi_{row}$ quality measure, for the \emph{Balance-scale} dataset ($\varphi_{row}(DS)=0.304$).}
\begin{tabular}{p{6cm} r r}
description & \#cases & $\varphi_{row}$ \\
\hline
$Left\_Weight \le 2 \land Right\_Distance \le 2 \land Right\_Weight > 3$ & 40 & 0.650 \\
$Left\_Weight \le 2 \land Right\_Distance \le 1 \land Right\_Weight > 2$ & 30 & 0.634 \\
$Left\_Weight \le 2 \land Right\_Distance \le 2 \land Right\_Weight > 2$ & 60 & 0.617 \\
$Left\_Weight \le 3 \land Right\_Distance \le 1 \land Right\_Weight > 3$ & 30 & 0.607 \\
$Left\_Weight \le 3 \land Right\_Distance \le 2 \land Right\_Weight > 3$ & 60 & 0.604 \\
$Left\_Distance > 3 \land Left\_Weight \le 2 \land Right\_Distance \le 2$ & 40 & 0.589 \\
\end{tabular}
\label{tab:balance_row}
\end{table}
 
% tab:balance_GT_yac
\begin{table}[t]
\centering
\caption{Subgroups found with the $\varphi_{GT\_yac}$ quality measure, for the \emph{Balance-scale} dataset ($\varphi_{GT\_yac}(DS)=0.310$).}
\begin{tabular}{p{6cm} r r}
description & \#cases & $\varphi_{GT\_yac}$ \\
\hline
$Left\_Weight \le 2 \land Right\_Distance \le 2 \land Right\_Weight > 3$ & 40 & 0.663 \\
$Left\_Weight \le 2 \land Right\_Distance \le 1 \land Right\_Weight > 2$ & 30 & 0.654 \\
$Left\_Weight \le 2 \land Right\_Distance \le 2 \land Right\_Weight > 2$ & 60 & 0.633 \\
$Left\_Distance > 2 \land Left\_Weight \le 1 \land Right\_Distance \le 2$ & 30 & 0.630 \\
$Left\_Distance > 3 \land Left\_Weight \le 1 \land Right\_Distance \le 3$ & 30 & 0.627 \\
$Left\_Weight \le 3 \land Right\_Distance \le 1 \land Right\_Weight > 3$ & 30 & 0.622 \\
\end{tabular}
\label{tab:balance_GT_yac}
\end{table}
 
% tab:balance_GT_yac2
\begin{table}[t]
\centering
\caption{Subgroups found with the $\varphi_{GT\_yac'}$ quality measure, for the \emph{Balance-scale} dataset ($\varphi_{GT\_yac'}=0.262$).}
\begin{tabular}{p{6cm} r r}
description & \#cases & $\varphi_{GT\_yac'}$ \\
\hline
$Left\_Distance > 2 \land Left\_Weight \le 1 \land Right\_Distance \le 2$ & 30 & 0.559 \\
$Left\_Distance > 3 \land Left\_Weight \le 1 \land Right\_Distance \le 3$ & 30 & 0.556 \\
$Left\_Weight \le 2 \land Right\_Distance \le 1 \land Right\_Weight > 2$ & 30 & 0.555 \\
$Left\_Weight \le 2 \land Right\_Distance \le 2 \land Right\_Weight > 3$ & 40 & 0.552 \\
$Left\_Weight \le 1 \land Right\_Distance \le 2 \land Right\_Weight > 1$ & 40 & 0.535 \\
$Left\_Weight \le 2 \land Right\_Distance \le 2 \land Right\_Weight > 2$ & 60 & 0.533 \\
\end{tabular}
\label{tab:balance_GT_yac2}
\end{table}

% img:balance:compare_row_GT_yac2
\begin{figure}[t]%
    \centering
    \subfloat[$Left\_Weight \le 2 \land Right\_Distance \le 2 \land Right\_Weight > 3$, top ranked by $\varphi_{row}$]{{\includegraphics[width=8cm]{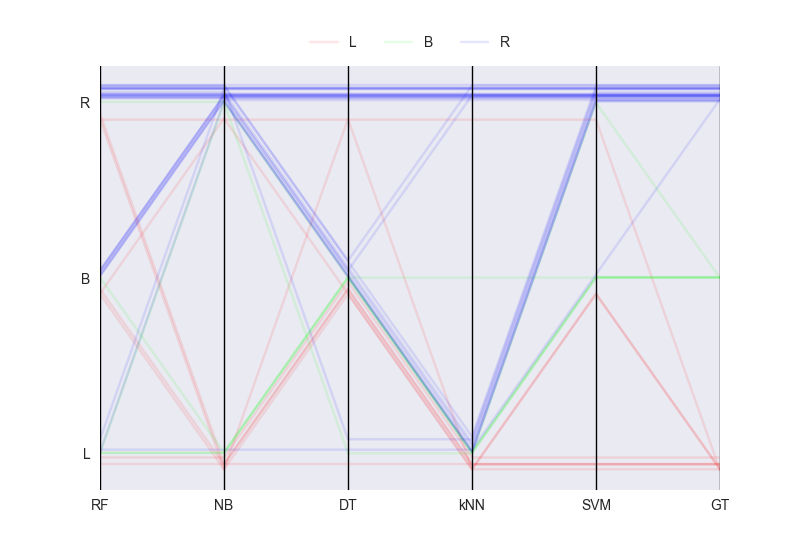}
    }}%
    \qquad
    \subfloat[$Left\_Distance > 2 \land Left\_Weight \le 1 \land Right\_Distance \le 2$, top ranked by $\varphi_{GT\_yac'}$]{{\includegraphics[width=8cm]{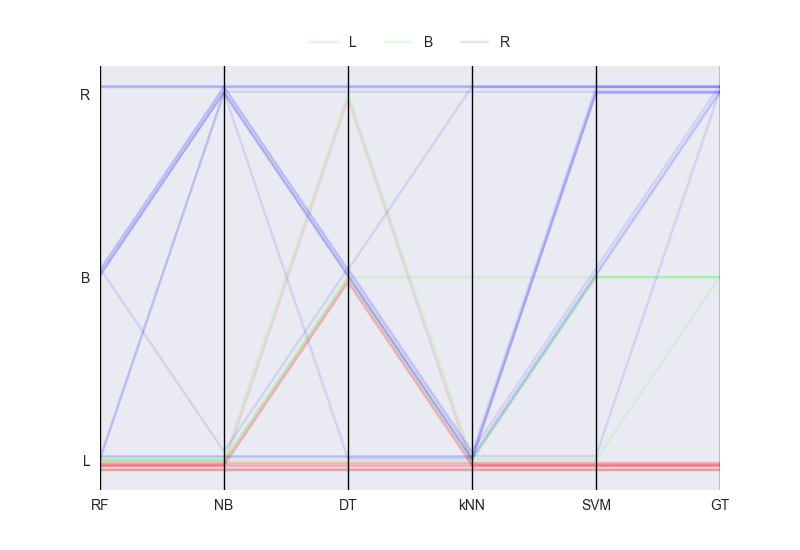}
    }}%
    \caption{Parallel coordinates. The \emph{Balance-scale} dataset.}
    \label{img:balance:compare_row_GT_yac2}
\end{figure}

% Balance fig_predictions_matrix_order=True_qmccl_1
\begin{figure}[t]
\centering
\includegraphics[width=8cm]{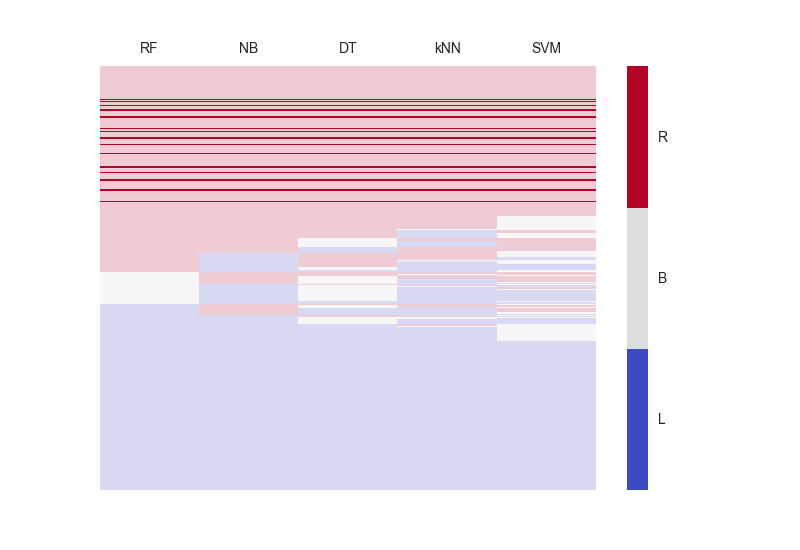}
\caption{Prediction matrix, \emph{Balance-scale} dataset, highlighted the subgroup for the description $Left\_Weight \le 3 \land Right\_Distance > 3 \land Right\_Weight > 4$, which is ranked highest by the $\varphi_{ccl}$ quality measure.}
\label{img:balance:fig_predictions_matrix_order=True_qmccl_1}
\end{figure}
 
%We next look into the \emph{Balance-scale} dataset.
The task for the Balance-scale dataset is classification, where 3 possible classes exist \emph{L} for left, \emph{B} for balanced, and \emph{R} for right.  The datasets represents a scale, where both on the left and on the right side a single weight is placed at a single spot.  For both the weight and the distance from the spot to the center of the balance, integer unit values between one and five can be chosen.  Hence, there are $5^4=625$ configurations.  The underlying physical law states that the scale is in balance, if and only if $Left\_Distance \times Left\_Weight$ equals $Right\_Distance \times Right\_Weight$. 

One can intuitively understand that, assuming the classifiers do not have access to the exact mechanism, higher confusion can be found near the decision boundaries and around the balanced state. 
%An example of the distribution of values and true labels is displayed in Figure~\ref{img:balance:fig_distribution_and_sg_qmrow_1}, in which a specific subgroup is highlighted in red.
%Figure~\ref{img:balance:fig_distribution_of_predictions} shows the distribution of predictions, and in Figure~\ref{img:balance:fig_predictions_correlation} we show heatmaps of the correlation among the classifiers' predictions (each heatmap is based on one-vs-all for a specific class). 
Na\"ive Bayes is expected to have some difficulties, as the assumption of independence among the conditional probabilities conflicts with the underlying multiplicative physical law (as just outlined). Naive Bayes resolves this problem by simply ignoring all the \emph{B} cases.  This is surprisingly effective, compared with the other classifiers.
A view of the predictions is displayed as a parallel coordinates plot in Figure~\ref{img:balance:fig_parallel_coordinates}, where the true labels are also included (far right).
%The order of the classifiers should bare no significance.
%The accuracy scores (on the whole dataset) are as follows, $92\%$ for the SVM, $88\%$ for the NB, $85\%$ for the kNN, $80\%$ for the RF, and $77\%$ for the DT. 

Table~\ref{tab:balance_row} lists the subgroups found with $\varphi_{row}$. The top description restricts three of the four variables: a small weight on the left side, and a large weight at a small distance on the right side.  This is indeed a volatile situation, where a small change in any of the remaining choices will cause the scales to tip over.  Hence, it makes sense that classifiers disagree.

Top subgroups for $\varphi_{GT\_yac}$ and $\varphi_{GT\_yac'}$ are listed in Tables \ref{tab:balance_GT_yac} and \ref{tab:balance_GT_yac2}, respectively.  This stepwise increase in the importance of the true label can be expected to affect the ranking of the top subgroups.  Indeed we see the description $Left\_Distance > 2 \land Left\_Weight \le 1 \land Right\_Distance \le 2$ appearing in the fourth place in $\varphi_{GT
\_yac}$, where it is not reported by $\varphi_{row}$, and then it climbs to the top in $\varphi_{GT\_yac'}$. 
%The effect of introducing the true label in additional columns, first increases the quality measure, in $\varphi_{GT\_yac}$ compared to $\varphi_{row}$, w.r.t.\ the whole dataset and the top reported descriptions, and then decreases it in $\varphi_{GT\_yac'}$.
In Figure~\ref{img:balance:compare_row_GT_yac2} we contrast the top subgroup for $\varphi_{row}$ (a) and the top subgroup for $\varphi_{GT\_yac'}$ (b). 

$\varphi_{ccl}$ reports descriptions for which the quality measure is 0. This value calculated from $0 - 0$, that is 0 for the mean per row entropies, and 0 for the per classifier entropy, for example $Left\_Weight \le 3 \land Right\_Distance > 3 \land Right\_Weight > 4$, which always results in \emph{R} true label, as can be seen in Figure~\ref{img:balance:fig_predictions_matrix_order=True_qmccl_1}. The $\varphi_{ccl}$ quality measure score for the whole dataset is $-0.942$. These descriptions correspond to a total agreement among the classifiers, which is not exactly what we are looking for yet is also interesting. To have descriptions for which the measure is bigger than 0, the mean per-row entropies should be higher than the mean per-classifier entropies. Similar reports are given by $\varphi_{cac}$ and by $\varphi_{cco}$.

$\varphi_{rasl}$ is not applicable here as we have more than two labels.   

%%%%%%%%%%%%%%%%%%%%%%%%%%%%%%%
% Titanic                     %
%%%%%%%%%%%%%%%%%%%%%%%%%%%%%%%

\subsection{Titanic}
\label{subsec:Titanic}

% tab:titanic_row
\begin{table}[t]
\centering
\caption{Subgroups found with the $\varphi_{row}$ quality measure, for the \emph{Titanic} dataset ($\varphi_{row}(DS)=0.294$).}
\begin{tabular}{p{6cm} r r}
description & \#cases & $\varphi_{row}$ \\
\hline
$Embarked = C \land Pclass \le 2 \land Sex = male$ & 52 & 0.842 \\
$Embarked \neq S \land Pclass \le 1 \land Sex = male$ & 43 & 0.838 \\
$Embarked = C \land Pclass \le 1 \land Sex = male$ & 42 & 0.835 \\
$Embarked = S \land Pclass > 2 \land Sex = female$ & 88 & 0.830 \\
$Age > 23 \land Pclass > 2 \land Sex = female$ & 43 & 0.804 \\
$Fare > 73.18988571428572 \land Sex = male$ & 40 & 0.780 \\
\end{tabular}
\label{tab:titanic_row}
\end{table}

% tab:titanic_GT_yac
\begin{table}[t]
\centering
\caption{Subgroups found with the $\varphi_{GT\_yac}$ quality measure, for the \emph{Titanic} dataset ($\varphi_{GT\_yac}(DS)=0.337$).}
\begin{tabular}{p{6cm} r r}
description & \#cases & $\varphi_{GT\_yac}$ \\
\hline
$Embarked \neq S \land Pclass \le 1 \land Sex = male$ & 43 & 0.849 \\
$Embarked = C \land Pclass \le 1 \land Sex = male$ & 42 & 0.846 \\
$Age > 23 \land Pclass > 2 \land Sex = female$ & 43 & 0.845 \\
$Embarked = C \land Pclass \le 2 \land Sex = male$ & 52 & 0.844 \\
$Embarked = S \land Pclass > 2 \land Sex = female$ & 88 & 0.837 \\
$Age > 11 \land Pclass > 2 \land Sex = female$ & 79 & 0.818 \\
\end{tabular}
\label{tab:titanic_GT_yac}
\end{table}

% tab:titanic_GT_yac2
\begin{table}[t]
\centering
\caption{Subgroups found with the $\varphi_{GT\_yac'}$ quality measure, for the \emph{Titanic} dataset ($\varphi_{GT\_yac'}(DS)=0.311$).}
\begin{tabular}{p{6cm} r r}
description & \#cases & $\varphi_{GT\_yac'}$ \\
\hline
$Age > 23 \land Pclass > 2 \land Sex = female$ & 43 & 0.750 \\
$Age > 11 \land Pclass > 2 \land Sex = female$ & 79 & 0.736 \\
$Embarked \neq S \land Pclass \le 1 \land Sex = male$ & 43 & 0.728 \\
$Embarked = C \land Pclass \le 1 \land Sex = male$ & 42 & 0.725 \\
$Embarked = C \land Pclass \le 2 \land Sex = male$ & 52 & 0.714 \\
$Embarked = S \land Pclass > 2 \land Sex = female$ & 88 & 0.711 \\
\end{tabular}
\label{tab:titanic_GT_yac2}
\end{table}

% img:titanic:compare_row_GT_yac2
\begin{figure}[t]%
    \centering
    \subfloat[$Embarked = C \land Pclass \le 2 \land Sex = male$, top ranked by $\varphi_{row}$]{{\includegraphics[width=8cm]{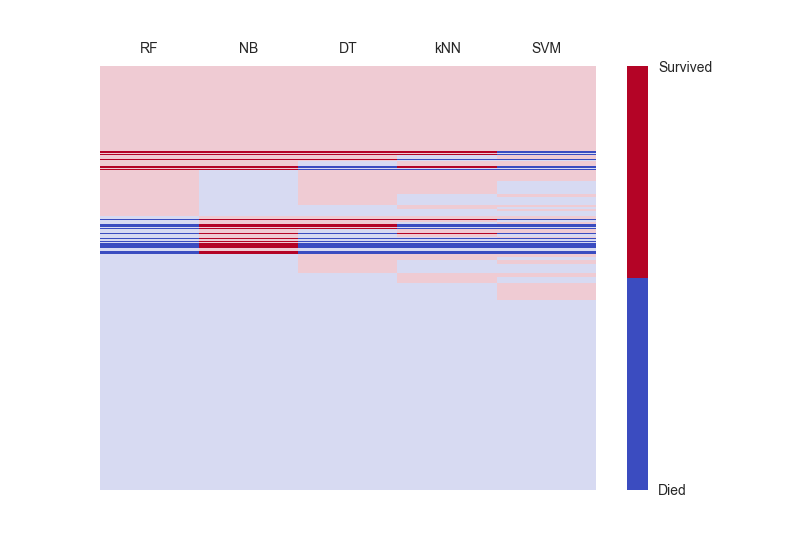}
    }}%
    \qquad
    \subfloat[$Age > 23 \land Pclass > 2 \land Sex = female$, top ranked by $\varphi_{GT\_yac'}$]{{\includegraphics[width=8cm]{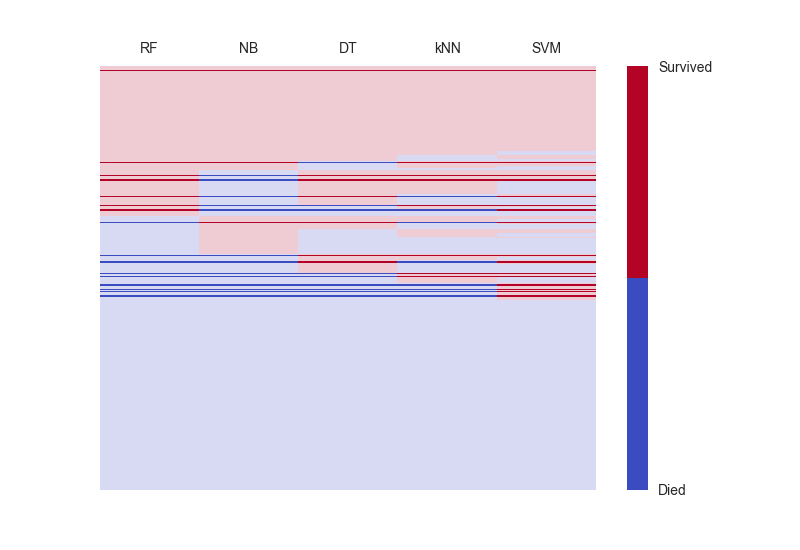} 
    }}%
    \caption{Prediction matrix ordered from left to right. The \emph{Titanic} dataset. For many cases highlighted for $\varphi_{row}$, Na\"ive Bayes is in a minority, predicting \emph{Survived}.}%
    \label{img:titanic:compare_row_GT_yac2}
\end{figure}

% img:titanic:compare_row_GT_yac2_pc
%\begin{figure}[t]%
%    \centering
%    \subfloat[$Embarked = C \land Pclass \le 2 \land Sex = male$, top ranked by $\varphi_{row}$]{{\includegraphics[width=8cm]{Images/Titanic/fig_parallel_coordinates_qmrow_1}
%    }}%
%    \qquad
%    \subfloat[$Age > 23 \land Pclass > 2 \land Sex = female$, top ranked by $\varphi_{GT\_yac'}$]{{\includegraphics[width=8cm]{Images/Titanic/fig_parallel_coordinates_qmGT_yac2_1} 
%    }}%
%    \caption{Parallel coordinates for the predictions and the true label. (a) The top description for $\varphi_{row}$. (b) The top description for $\varphi_{GT\_yac'}$. The \emph{Titanic} dataset. For significant number of the cases in (a), Na\"ive Bayes is right in predicting \emph{Survived}.}%
%    \label{img:titanic:compare_row_GT_yac2_pc}
%\end{figure}

% tab:titanic_rasl_max
\begin{table}[t]
\centering
\caption{Subgroups found with the $\varphi_{GT\_rasl}$ quality measure, for the \emph{Titanic} dataset ($\varphi_{rasl}(DS)=274.0$).}
\begin{tabular}{p{6cm} r r}
description & \#cases & $\varphi_{rasl}$ \\
\hline
$Parch \le 0 \land Pclass > 1 \land Sex = male$ & 385 & 127.32 \\
$Parch \le 1 \land Pclass > 1 \land Sex = male$ & 429 & 105.93 \\
$Pclass > 1 \land Sex = male$ & 455 & 105.74 \\
$Fare \le 146.38 \land Pclass > 1 \land Sex = male$ & 455 & 105.74 \\
$Fare \le 292.76 \land Pclass > 1 \land Sex = male$ & 455 & 105.74 \\
$Fare \le 365.95 \land Pclass > 1 \land Sex = male$ & 455 & 105.74 \\
\end{tabular}
\label{tab:titanic_rasl_max}
\end{table}

% tab:titanic_rasl_min
\begin{table}[t]
\centering
\caption{Subgroups found when minimizing the $\varphi_{rasl}$ quality measure, for the \emph{Titanic} dataset. $Age \le -1$ is referring to cases for which the age is not recorded. The classifiers used the median age for those cases.}
\begin{tabular}{p{6cm} r r}
description & \#cases & $\varphi_{rasl}$ \\
\hline
$Age \le -1$ & 177 & 0.0 \\
$Age \le -1 \land Pclass > 1$ & 147 & 0.0 \\
$Age \le -1 \land Pclass \le 2$ & 41 & 0.0 \\
$Age \le -1 \land Pclass > 2$ & 136 & 0.0 \\
$Age \le -1 \land Sex = male$ & 124 & 0.0 \\
$Age \le -1 \land Sex = female$ & 53 & 0.0 \\
\hline
\end{tabular}
\label{tab:titanic_rasl_min}
\end{table}

% Titanic fig_predictions_matrix_order=True_qmrasl_max_1
\begin{figure}[t]
\centering
\includegraphics[width=8cm]{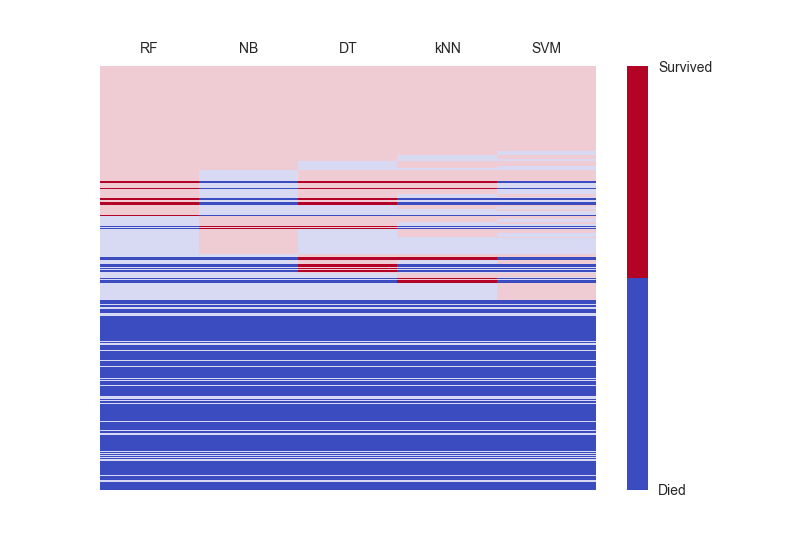}
\caption{$Parch \le 0 \land Pclass > 1 \land Sex = male$, \emph{Titanic} dataset. This description is ranked at the top when maximizing $\varphi_{rasl}$. SVM does the best by predicting always \emph{Died}.}
\label{img:titanic:fig_predictions_matrix_order=True_qmrasl_max_1}
\end{figure}

% tab:titanic_ccl
\begin{table}[t]
\centering
\caption{Subgroups found with the $\varphi_{ccl}$ quality measure, for the \emph{Titanic} dataset ($\varphi_{ccl}(DS)=-0.645$).}
\begin{tabular}{p{6cm} r r}
description & \#cases & $\varphi_{ccl}$ \\
\hline
$Embarked = C \land Pclass \le 2 \land Sex = male$ & 52 & 0.313 \\
$Embarked \neq S \land Pclass \le 1 \land Sex = male$ & 43 & 0.291 \\
$Embarked = C \land Pclass \le 1 \land Sex = male$ & 42 & 0.290 \\
$Embarked \neq S \land Pclass \le 2 \land Sex = male$ & 54 & 0.284 \\
$Embarked = S \land Pclass > 2 \land Sex = female$ & 88 & 0.205 \\
$Age > 23 \land Pclass > 2 \land Sex = female$& 43 & 0.097 \\
\end{tabular}
\label{tab:titanic_ccl}
\end{table}

% tab:titanic_cac
\begin{table}[t]
\centering
\caption{Subgroups found with the $\varphi_{cac}$ quality measure, for the \emph{Titanic} dataset ($\varphi_{cac}(DS)=-0.178$).}
\begin{tabular}{p{6cm} r r}
description & \#cases & $\varphi_{cac}$ \\
\hline
$Embarked = C \land Pclass \le 1 \land Sex = male$ & 42 & 0.116 \\
$Embarked \neq S \land Pclass \le 1 \land Sex = male$ & 43 & 0.103 \\
$Age \le 22 \land Embarked = S \land SibSp > 2$ & 38 & 0.031 \\
$Age \le 22 \land Embarked \neq Q \land SibSp > 2$ & 38 & 0.031 \\
$Embarked = S \land Pclass > 2 \land SibSp > 2$ & 38 & 0.021 \\
$Embarked \neq Q \land Pclass > 2 \land SibSp > 2$ & 38 & 0.021 \\
\end{tabular}
\label{tab:titanic_cac}
\end{table}

% tab:titanic_cco
\begin{table}[t]
\centering
\caption{Subgroups found with the $\varphi_{cco}$ quality measure, for the \emph{Titanic} dataset ($\varphi_{cco}(DS)=-0.457$).}
\begin{tabular}{p{6cm} r r}
description & \#cases & $\varphi_{cco}$ \\
\hline
$Fare > 73.19 \land Parch \le 1 \land Sex = female$ & 52 & -0.014 \\
$Fare > 73.19 \land Parch \le 0 \land Sex = female$ & 41 & -0.015 \\
$Fare \le 146.38 \land Fare > 73.19 \land Sex = female$ & 41 & -0.031 \\
$Parch \le 1 \land Pclass \le 1 \land Sex = female$ & 81 & -0.110 \\
$Fare \le 146.38 \land Pclass \le 1 \land Sex = female$ & 73 & -0.124 \\
$Parch \le 0 \land Pclass \le 1 \land Sex = female$ & 64 & -0.131 \\
\end{tabular}
\label{tab:titanic_cco}
\end{table}

% Titanic fig_predictions_matrix_order=True_qmcac_3
\begin{figure}[t]
\centering
\includegraphics[width=8cm]{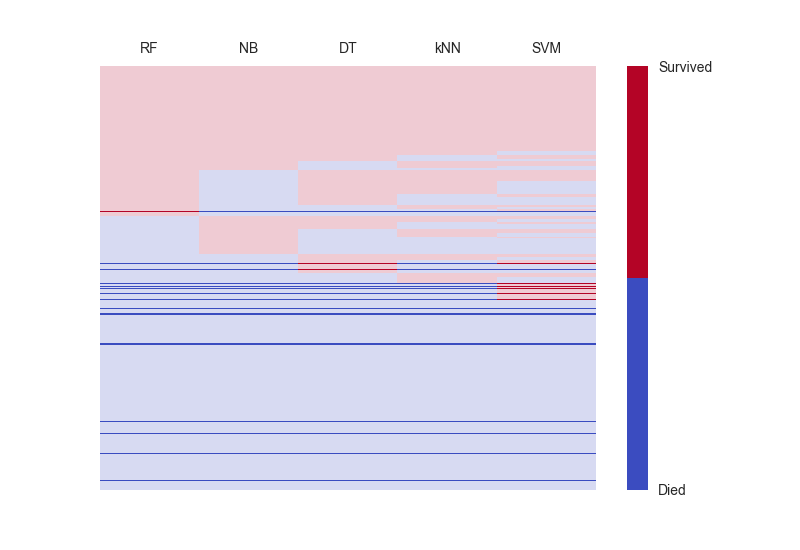}
\caption{$Age \le 22 \land Embarked = S \land SibSp > 2$, \emph{Titanic} dataset. This is the third description reported by $\varphi_{cac}$. The accordance with the majority, leads to apparent internal consistency. Yet the majority of classifiers may be wrong for some of the cases in this region, and therefore this description is not be ranked high for $\varphi_{cco}$.}
\label{img:titanic:fig_predictions_matrix_order=True_qmcac_3}
\end{figure}

The task for the \emph{Titanic} dataset is binary classification (life or death), based on attributes known about the passengers of the Titanic. The famous ship collected passengers from three ports. Some of the passengers traveled alone, while others traveled with family members. There were three classes of cabinets with different price levels. In general, once the ship hit the iceberg, children and women were offered a place in a lifeboat before the other passengers. Unfortunately there were not enough boats for everyone.

Table~\ref{tab:titanic_row}, \ref{tab:titanic_GT_yac}, and \ref{tab:titanic_GT_yac2} list the top subgroups found with $\varphi_{row}$, $\varphi_{GT\_yac}$, and $\varphi_{GT\_yac'}$, respectively. 
As can be seen in Figure~\ref{img:titanic:compare_row_GT_yac2}, the top description for $\varphi_{row}$ picks on a region where the Na\"ive Bayes classifier predicts mostly \emph{Survived}. This is also the case for the top description for $\varphi_{GT\_yac}$, which is ranked in the second place for $\varphi_{row}$. By contrast, $\varphi_{GT\_yac'}$ ranks other descriptions first.
%We also show that in a parallel coordinates plot. Note that in Figure~\ref{img:titanic:compare_row_GT_yac2_pc} the rightmost tick is the true label.

Tables~\ref{tab:titanic_rasl_max} and \ref{tab:titanic_rasl_min} list subgroups found when maximizing and minimizing, respectively, $\varphi_{rasl}$.  The top description from maximizing $\varphi_{rasl}$ is $Parch \le 0 \land Pclass > 1 \land Sex = male$. SVM has the highest accuracy $89\%$ in this region by predicting all those 385 cases as Died, while 44 passengers did survive (cf.\@ Figure~\ref{img:titanic:fig_predictions_matrix_order=True_qmrasl_max_1}). All attempts by the other classifiers to identify the survivors result in many false positives. The top description when minimizing $\varphi_{rasl}$, $Age \le -1$, refer to cases for which the age is unknown.
The Decision Tree and the Random Forest classifiers, achieve in this region the top accuracy of $85\%$, by predicting correctly 39 \emph{Survived} cases out of 52. Other models also predict  correctly most of the positive cases.
We have maximized $\varphi_{rasl}$ and found a region for which the SVM classifier achieves the highest accuracy of $89\%$, and we have minimized $\varphi_{rasl}$ and found a region for which those are the DT and RF classifiers that achieve the highest accuracy, this time only $85\%$, which is lower than $89\%$. Therefore we note that $\varphi_{rasl}$ with this setting of classifiers' predictions, is not discriminating subgroups based on individual classifiers' accuracy but rather based on whether the collection of classifiers, if used as a voting ensemble, correctly ranks the cases. Therefore, as a voting ensemble the classifiers do a better job for $Age \le -1$ than for $Parch \le 0 \land Pclass > 1 \land Sex = male$.

Tables~\ref{tab:titanic_ccl}, \ref{tab:titanic_cac}, and \ref{tab:titanic_cco} list the top subgroups for $\varphi_{ccl}$, $\varphi_{cac}$, and $\varphi_{cco}$, respectively. Subgroups for $\varphi_{ccl}$ are very similar to those for $\varphi_{row}$. The top two subgroups from $\varphi_{cac}$ are also found with $\varphi_{row}$ and with $\varphi_{cco}$. Those subgroups correspond to regions in which there is internal agreement for NB and for SVM, as in Figure~\ref{img:titanic:compare_row_GT_yac2} (a). The third ranked description in $\varphi_{cac}$ is given in Figure~\ref{img:titanic:fig_predictions_matrix_order=True_qmcac_3}. The fact that there are differences between $\varphi_{cac}$ and $\varphi_{cco}$ tells us that sometimes the classifiers agree, yet they agree on the wrong prediction. The top ranked subgroups for $\varphi_{cco}$ are of regions with a lot of agreement for \emph{Survived}.

\subsection{YearPredictionMSD}

% Year fig_distribution_of_predictions
%\begin{figure}[t]
%\centering
%\includegraphics[width=8cm]{Images/YearPredictionMSD/fig_distribution_of_predictions}
%\caption{The \emph{YearPredictionMSD} dataset. Distribution of predictions.}
%\label{img:year:fig_distribution_of_predictions}
%\end{figure}

% tab:year_row
\begin{table}[t]
\centering
\caption{Subgroups found with the $\varphi_{row}$ quality measure, for the \emph{YearPredictionMSD} dataset ($\varphi_{row}(DS)=1.07597$).}
\begin{tabular}{p{6cm} r r}
description & \#cases & $\varphi_{row}$ \\
\hline
$fea13 \le 78.65 \land fea3 > 55.48 \land fea6 > -10.93$ & 21375 & 1.42877 \\
$fea3 > 55.48 \land fea6 > -10.93 \land fea60 > -159.88$ & 21414 & 1.42746 \\
$fea3 > 55.48 \land fea6 > -10.93 \land fea72 > -270.76$ & 21090 & 1.42742 \\
$fea1 \le 44.7641 \land fea13 \le 78.65 \land fea3 > 55.48$ & 22172 & 1.42707 \\
$fea10 > -12.50 \land fea3 > 55.48 \land fea6 > -10.93$ & 21773 & 1.42465 \\
$fea3 > 55.48 \land fea6 > -10.93 \land fea7 \le 17.85$ & 21158 & 1.42462 \\
\end{tabular}
\label{tab:year_row}
\end{table}
 
% tab:year_ccl
\begin{table}[t]
\centering
\caption{Subgroups found with the $\varphi_{ccl}$ quality measure, for the \emph{YearPredictionMSD} dataset ($\varphi_{ccl}(DS)=-0.48005$).}
\begin{tabular}{p{6cm} r r}
description & \#cases & $\varphi_{ccl}$ \\
\hline
$fea1 > 44.76 \land fea38 \le 278.49 \land fea6 \le -10.93$ & 129,197 & 0.03053 \\
$fea1 > 44.76 \land fea56 \le 35.74 \land fea6 \le -10.93$ & 112,228 & 0.02556 \\
$fea1 > 44.76 \land fea6 \le -10.93 \land fea70 \le 8.27$ & 105,874 & 0.02076 \\
$fea1 > 44.76 \land fea6 \le -10.93 \land fea65 > -51.89$ & 110,647 & 0.01772 \\
$fea1 > 44.76 \land fea10 > 2.07 \land fea6 \le -10.93$ & 93,033 & 0.01620 \\
$fea1 > 44.76 \land fea11 \le -2.09 \land fea6 \le -10.93$ & 30,640 & 0.01467 \\
\end{tabular}
\label{tab:year_ccl}
\end{table}

On the YearPredictionMSD dataset we define a classification task with 10 classes. The features are real numbers, taken from the Echo Nest API, for analyzing sound tracks. The first 12 features are averages of timbre values (fea1-fea12), and the next 78 features are covariances of timbre values (fea13-fea90).
It is assumed that from the sound analysis of the songs it is possible in theory, to predict the year the song was released (or in our case the decade).
%The distribution of prediction is brought in Figure~\ref{img:year:fig_distribution_of_predictions}.
The accuracies that our classifiers reported are as follows: $61\%$ for SVM, $58\%$ for RF, $51\%$ for kNN, $48\%$ for DT, and $23\%$ for NB.

Tables~\ref{tab:year_row} and \ref{tab:year_ccl} list subgroups found with $\varphi_{row}$ and $\varphi_{ccl}$, respectively. The latter delivers larger subgroups (cf.\@ Table~\ref{tab:year_ccl}).
For both $\varphi_{cac}$ and $\varphi_{cco}$, the description $fea1 > 44.76 \land fea11 \le -2.09 \land fea6 \le -10.93$, appears at the top, while for $\varphi_{ccl}$, as can be seen in Table~\ref{tab:year_ccl}, that description appears last.
%For $\varphi_{GT\_yac}$ and $\varphi_{GT\_yac'}$ we see here also the phenomena, w.r.t.\ the transition from $\varphi_{row}$ to $\varphi_{GT\_yac}$ and then to $\varphi_{GT\_yac'}$, where new descriptions climb and push existing descriptions down.

%%%%%%%%%%%%%%%%%%%%%%%%%%
% Car, Pima-indians, Adult
%%%%%%%%%%%%%%%%%%%%%%%%%%

\subsection{Covertype, Car, Pima-indians, Adult}

We find similar observations with \emph{Covertype}, \emph{Car}, \emph{Pima-indians}, and with \emph{Adult}.  Therefore, to save space, we will not list the subgroups found on these datasets here.  
$\varphi_{row}$ reports regions with high controversy among classifiers.
Moving to $\varphi_{GT\_yac}$ and then to $\varphi_{GT\_yac'}$ there is a trend of new descriptions climbing in the ranking and pushing other descriptions down, as more emphasis is given to cases predicted wrongly.

For $\varphi_{ccl}$, $\varphi_{cac}$, and $\varphi_{cco}$ the top descriptions are sometimes regions with a lot of agreement, or in more interesting situations, regions with high controversy, similar to those reported by $\varphi_{row}$, yet of more structured nature, where a few classifiers differ consistently from the rest.
In an informal manner, we can say, that for \emph{Adult}, $\varphi_{ccl} \neq \varphi_{cac} = \varphi_{cco}$,
and for \emph{Car}, $\varphi_{ccl} = \varphi_{cco} \neq \varphi_{cac}$.
This difference has to do with whether the majority of classifiers predict correctly or wrongly, in a significant trend, in a region.

$\varphi_{rasl}$ when maximized for \emph{Adult} gives different and bigger subgroups than those reported by $\varphi_{row}$, and this is the case also for \emph{Pima-indians}. Minimizing $\varphi_{rasl}$ gives well defined regions %w.r.t.\ the classification task, 
for which most classifiers predict correctly.

\section{Conclusions}
\label{sec:conclusion}

We introduce the Controversy Rules model class for Exceptional Model Mining, to find regions of the input space where a set of classifiers is in unusual (dis-)agreement.  This level of (dis-)agreement can be gauged in many different ways; we introduce several quality measures to explore various options. $\varphi_{row}$ reports regions with high controversy among the classifiers.
$\varphi_{GT\_yac}$ is similar to $\varphi_{row}$ yet puts some emphasis also on errors. $\varphi_{GT\_yac'}$ puts even more emphasis on errors. We demonstrate, in Figure~\ref{img:titanic:compare_row_GT_yac2} this trend.
Comparing descriptions reported by $\varphi_{ccl}$ to those reported by $\varphi_{row}$ tells us how consistent the controversy is, and whether internal consistency consideration yields different descriptions, as shown in Figure~\ref{img:mushroom:compare_row_ccl}.
Further comparing to $\varphi_{cac}$ and to $\varphi_{cco}$, shows us also regions where classifiers have different predictions, and whether the majority of classifiers predict there correctly, as we show in Figure~\ref{img:mushroom:fig_predictions_matrix_order=True_qmcac_1} and in Figure~\ref{img:titanic:fig_predictions_matrix_order=True_qmcac_3}. Occasionally the reported subgroups with $\varphi_{ccl}$, $\varphi_{cac}$, and/or $\varphi_{cco}$, correspond to regions in which there is a lot of agreement among the classifiers, as for other regions, on average, the per classifier internal controversy is higher than the averaged per row controversy.
$\varphi_{rasl}$ is treating the collection of classifiers as a single voting-based ensemble, and shows us when this ensemble predicts (in-)correctly, as is discussed over the \emph{Titanic} dataset in Subsection~\ref{subsec:Titanic}.
For the \emph{Mushroom} dataset, in Subsection~\ref{subsec:mushroom} we see a setting where $\varphi_{row}$ reports interesting regions, while $\varphi_{rasl}$ does not identify anything special.
It is not surprising that there are similarities between the reports of the various quality measures, yet all quality measures evaluated in this paper are useful and shed unique light. The differences among the reports, give us additional clues to better understand the classifiers and their interaction with the modalities in the data.

We note the following challenges. All above quality measures are designed under the assumption, that all predictions are available (matrix $M$ is full); missing values cannot be accommodated. 
%After the first analysis, with a given set of classifiers, and with no additional restrictions, one may opt to take out some of the classifiers, or to black-list specific regions.
%At the moment, one can do that trivially, yet manually.
While our experiments are with some unbalanced datasets, we do not study in detail the effects of this.
We account for all discrepancies the same, yet with ordinal classes, some differential weight is more appropriate. 
Finally, an appropriate adjustment for regression tasks is nontrivial. 

For future work we see two promising directions. The one is exploring ensembles. The differences between what we are doing in this paper and studying ensembles, is that we evaluate few classifiers of different type, while for ensembles there usually is a bigger number of base estimators, yet of more similar nature. The other promising direction we identify is focusing the discussion on a specific classifier's interpretability, contrasting the model with possible explanations.

\end{document}